\documentclass[onecolumn]{amd}
\usepackage{graphicx}
\usepackage{subcaption,ragged2e}
\usepackage[numbers,round]{natbib}
\usepackage[normalem]{ulem}
\usepackage[utf8]{inputenc}
\usepackage[T1]{fontenc}
\usepackage{url}
\usepackage{tikz}
\usepackage{enumitem}
\usetikzlibrary{positioning}
\usetikzlibrary{calc}
\usepackage{array}
\usepackage{booktabs, multirow}
\usepackage{wrapfig}
\usepackage{caption}
\usepackage{hyperref}
\usepackage{listings}
\usepackage{adjustbox}
\usepackage{footnote}
\usepackage{comment}
\usepackage{multirow}
\usepackage{amsmath}
\usepackage{amsfonts}
\usepackage{breakcites}
\usepackage{blindtext}
\usepackage[most]{tcolorbox}
\usepackage{tabularx}
\usepackage{graphicx} 
\usepackage{xcolor}
\definecolor{persimmon}{RGB}{236,88,0}

\title{MXFP4 for LLM Inference: Complementary Post-Training Quantization Techniques and Their Interactions}
\author{Natalia Frumkin\IEEEauthorrefmark{1}, Alireza Khodamoradi, Kristof Denolf, Eric Dellinger, Stuart Biles\\AMD Research \thanks{\IEEEauthorrefmark{1}Correspondence to:
  \texttt{nfrumkin@amd.com}.}
}
\date{}

\makeatletter
\def\paragraph{\@startsection{paragraph}{4}{2\parindent}{0ex plus 0.1ex minus 0.1ex}%
{0ex}{\normalfont\normalsize\bfseries}}%

\def\@IEEEsectpunct{.\ }%
\makeatother

\usepackage{etoolbox}
\makeatletter
\patchcmd{\@maketitle}{\newpage}{\newpage\vspace*{0.15in}}{}{}
\makeatother

\begin{document}

\maketitle

\begin{tikzpicture}[remember picture, overlay]
  \node[anchor=north west, inner sep=0pt]
       at ([xshift=0.75in, yshift=-0.5in]current page.north west)
       {\includegraphics[width=110pt]{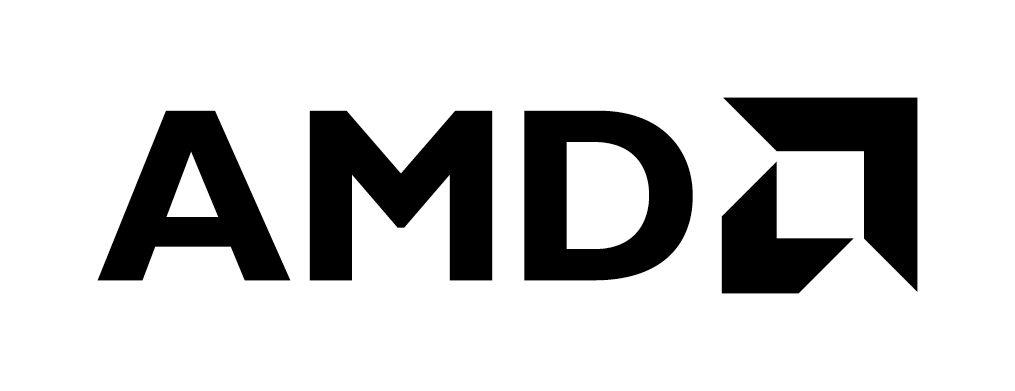}};
\end{tikzpicture}\hspace*{\fill}

\begin{abstract}
Recent advances in low-precision representation have introduced several microscaling numerical formats for LLM quantization. By sharing a scale factor across groups of 16 or 32 elements, these formats enable aggressive low-precision quantization with minimal accuracy degradation, particularly when combined with modern PTQ algorithms. Their accuracy depends on how the numerical format interacts with rotation, PTQ, and scale selection. We study the composability of PTQ configurations across six LLMs and three microscaling formats: MXFP6, MXFP4, and NVFP4, systematically varying Hadamard rotation granularity, PTQ algorithm, and scale-selection strategy. Our central finding is that effective quantization recipes are format dependent: both MXFP4 and NVFP4 support accurate 4-bit inference, but techniques that benefit one do not necessarily transfer to the other. MXFP4 benefits substantially from rotation and error-correcting PTQ, whereas NVFP4 gains less from rotation, consistent with its finer-grained scales capturing more within-block variation.
Building on these findings, we introduce \textbf{Midmax}, a calibration-free scale-selection strategy that mitigates clipping error in power-of-two block scales and improves accuracy across a majority of the evaluated MXFP4 configurations. We show that pairing MXFP6 activations with MXFP4 weights recovers most of the accuracy lost at 4-bit and, on Llama-3.1-8B, matches or exceeds our in-house BF16 baseline. An additional study finds that augmenting MXFP4 with a per-tensor macroscale, a mechanism intrinsic to NVFP4, does not improve average accuracy, suggesting that global rescaling alone does not address the limitations of E8M0 block scales for MXFP4 quantization. Together, these results provide practical guidance for combining numerical formats, rotation, PTQ, and scale selection for accurate low-precision LLM inference.
\end{abstract}



\section{Introduction}

Memory bandwidth, energy consumption, and hardware efficiency are primary system-level constraints in neural network inference, from single-device deployments to large-scale serving systems~\citep{firoozshahian2023mtia,stojkovic2024towards}. Modern large language models (LLMs) are dominated by dense linear algebra and memory movement, and since memory traffic and compute cost scale directly with operand precision, numerical format is a central system-level decision rather than a low-level implementation detail~\citep{dao2022flashattention,zheng2024sglang}.
The adoption of FP8 in recent hardware has demonstrated that low-precision floating-point formats can significantly improve throughput and energy efficiency while maintaining model quality~\citep{micikevicius2022fp8}. Motivated by these gains, prior work has pushed to 4-bit precision, with post-training quantization (PTQ) techniques such as weight rounding optimization~\citep{frantar2022gptq}, activation outlier smoothing~\citep{xiao2023smoothquant}, and sequential error correction~\citep{zhang2025qronoscorrectingpastshaping} establishing that transformer-based LLMs can maintain accuracy at 4-bit.
Block floating-point (BFP) representations have emerged as a natural fit for 4-bit LLM inference. Rather than applying a single scale globally, BFP shares a scale across a local block of elements, providing per-block dynamic range coverage while remaining compatible with efficient hardware implementation. The OCP Microscaling (MX) specification has standardized this approach~\citep{OCP-MX-v1.0}, and recent hardware including AMD Instinct accelerators with the CDNA4 architecture provide native support for MXFP4 arithmetic~\citep{amd-cdna4-isa}, making 4-bit block-scaled inference a practical deployment target.

We study three microscaling formats in this work. MXFP6 and MXFP4 are defined by the OCP standard and use a power-of-two (E8M0) block scale shared across 32 elements, with MXFP6 providing a higher-precision 6-bit element datatype and MXFP4 a 4-bit one. NVFP4 instead uses a finer-grained FP8 (E4M3) block scale shared across 16 elements, together with an additional high-precision per-tensor macroscale. On Llama-3 and Qwen3 models evaluated with standard zero-shot and math benchmarks, published results show NVFP4 ahead of MXFP4 when both 4-bit formats are paired with the same off-the-shelf PTQ method~\citep{arxiv2509.23202}; concurrent work on Llama-3.1-8B-Instruct and Qwen3-8B shows that augmenting MXFP4 with specialized scale-factor techniques can close this gap to within 1\% of NVFP4 on the same benchmarks~\citep{arxiv2603.08713}. These comparisons, however, typically pair each format with a single fixed set of complementary techniques, which makes it difficult to tell how much of the observed behavior comes from the format and how much from the Hadamard rotation, PTQ algorithm, and scale-selection strategy applied alongside it.

In this work, we instead study the \emph{composability} of PTQ techniques with microscaling formats: for each format we vary Hadamard rotation granularity, PTQ algorithm, and scale-selection strategy across six LLMs, and measure how these techniques combine. Our central finding is that, with an appropriate recipe, both MXFP4 and NVFP4 are strong candidates for 4-bit (A4W4) inference, and that MXFP6 recovers most of the remaining accuracy at 6-bit; the techniques that compose well with each format, however, differ. For MXFP4 at A4W4, Midmax scale selection, block-level Hadamard rotation, and PTQ yield approximately $3\%$ cumulative gain over baseline, whereas NVFP4 benefits from its intrinsic macroscale and smaller block size, as well as from error-correcting PTQ algorithms. The right post-training quantization choice is therefore a complete format-plus-technique recipe.

\paragraph{Contributions}
This work provides a study of how PTQ techniques compose with microscaling formats across six LLMs and makes the following contributions:

\begin{itemize}

\item \textbf{A composability study of PTQ techniques across microscaling formats.}
We measure how Hadamard rotation, PTQ algorithm, and scale selection combine with MXFP6, MXFP4, and NVFP4. With appropriate techniques both MXFP4 and NVFP4 are viable at A4W4, but the effective recipe differs by format: MXFP4 benefits strongly from block-level rotation and error-correcting PTQ algorithms, while the same rotation is often redundant or harmful for NVFP4. We further show that block Hadamard rotation yields better performance than full-sized rotation at A4W4.

\item \textbf{Midmax provides effective, calibration-free scale selection.}
We introduce Midmax, which shifts the rounding boundary of power-of-two scales to reduce clipping error. It improves upon floor rounding in 67\% of the evaluated A4W4 configurations, without requiring calibration data.

\item \textbf{Mixed-precision MXFP6/MXFP4 and a macroscale study.}
We show that assigning MXFP6 to activations and MXFP4 to weights (A6W4) is an effective mixed-precision trade-off, matching or exceeding our in-house BF16 baseline. As a secondary study, we also examine whether NVFP4's per-tensor macroscale benefits MXFP4; although it yields occasional gains on individual tasks and configurations, it does not improve accuracy on average, indicating that an effective MXFP4 recipe does not require macroscaling.

\end{itemize}

\section{Background: Block Floating-Point and Complementary Techniques}

This section establishes the quantization framework and BFP representations that underpin our study. We describe symmetric quantization, BFP formats, and the various PTQ configuration choices summarized in Table~\ref{tab:bfp_designs}.

\subsection{Quantization and Dequantization}

We consider symmetric quantization, which maps floating point values symmetrically around zero. To convert a floating-point value $x$ to its quantized representation, we apply the following mapping:
\begin{equation}
    x_q = \operatorname{round}_{\mathcal{F}}\!\left(\frac{x}{\Delta}\right)
    \label{eq:quant}
\end{equation}

where $x_q$ is the quantized value, $\Delta$ is the quantization scale, and the $\operatorname{round}_{\mathcal{F}}$ operator rounds its input to a representable value in the target floating-point alphabet. For FP4's E2M1 datatype, the target alphabet is $\{\pm0, \pm0.5, \pm1.0, \pm1.5, \pm2.0, \pm3.0, \pm4.0, \pm6.0\}$. The reconstructed floating-point value $\hat{x}$ is recovered via dequantization:
\begin{equation}
    \hat{x} = \Delta \cdot x_q.
    \label{eq:dequant}
\end{equation}
In practice, not all model components are quantized uniformly; non-linear layers and embedding tables are typically retained in higher precision (e.g., BF16) due to their sensitivity to quantization noise~\citep{frantar2022gptq, dettmers2022gpt3}.

\subsection{Block Floating-Point Representations}

In high-precision floating point, each value carries its own private exponent, providing fine-grained per-element dynamic range at the cost of storing one high-precision exponent per value, for example, an 8-bit exponent in FP32 and BF16 datatypes. Block floating-point (BFP) representations reduce per-element dynamic range and compensate with a single shared scale $\Delta$ across a block of $k$ elements, amortizing the block scale cost while maintaining local dynamic range coverage, as in MXFP4's E2M1, which pairs a 2-bit element exponent with a shared block scale. The OCP Microscaling (MX) specification~\citep{OCP-MX-v1.0} has standardized this approach, and recent AMD Instinct accelerators with the CDNA4 architecture provide native hardware support~\citep{amd-cdna4-isa}.

Within the OCP MX family, three configuration axes determine numerical behavior: the \emph{element datatype}, the \emph{scale datatype}, and the \emph{block layout}. In contrast, NVFP4 adds another axis with an additional per-tensor macroscale; it also has a different block scale datatype and a smaller block size. The macroscale in NVFP4 is software-implemented and is not an exclusive hardware feature; it can therefore be applied to other block-scaled formats. In this work, we also study MXFP4 augmented with a per-tensor macroscale. Table~\ref{tab:bfp_designs} summarizes the four block-scaled configurations in this study.

\begin{table}[h]
    \centering
    \small
    \setlength{\tabcolsep}{7pt}
    \renewcommand{\arraystretch}{1.2}
    \begin{tabular}{lcccc}
        \toprule
        Format & Element Datatype ($x_q$) & Block Layout ($1 \times k$) & Scale Datatype ($\Delta$) & Macroscale \\
        \midrule
        MXFP4              & E2M1 & $1 \times 32$ & E8M0 & None \\
        MXFP4+Macroscale   & E2M1 & $1 \times 32$ & E8M0 & BF16 \\
        NVFP4              & E2M1 & $1 \times 16$ & E4M3 & BF16 \\
        \midrule
        MXFP6 & E2M3 & $1 \times 32$ & E8M0 & None \\
        \bottomrule
    \end{tabular}

    \caption{The 4-bit and 6-bit Block Floating-Point datatype formats compared in this work. MXFP4 and MXFP4+macroscale share the same element datatype and block layout but differ in the presence of a per-tensor macroscale. NVFP4 uses finer-grained E4M3 block scales, smaller blocks, and a per-tensor macroscale.}
    \label{tab:bfp_designs}
\end{table}

\paragraph{Scale computation}
The block scale $\Delta$ is a function of the $k$ elements in the block. The OCP MX standard~\citep{OCP-MX-v1.0} specifies a power-of-two (E8M0) datatype for the block scale $\Delta$. OCP MX's companion paper~\citep{rouhani2023microscaling} describes the scale calculation for a block of elements, $\{x_1, \ldots, x_{32}\}$, in two steps: first, the block exponent $E_{\mathrm{block}}$ is set to the floor of the log of the block maximum,
\begin{equation}
    E_{\mathrm{block}} = \left\lfloor \log_2 \left( \max_{i=1}^{32} |x_i| \right) \right\rfloor;
\end{equation}
then the scale is obtained by shifting this exponent down by the element datatype's largest representable exponent, $e_{\max}^{\mathrm{elem}}$ (for FP4 E2M1, $e_{\max}^{\mathrm{elem}} = 2$):
\begin{equation}
    \label{eq:sharedScale}
    \Delta = 2^{E_{\mathrm{block}} - e_{\max}^{\mathrm{elem}}}.
\end{equation}
This rescales values relative to the largest magnitude in the block before rounding and clipping. Since each block uses its own scale derived from local statistics, block-scaled representations can handle large variations in value range across blocks. This is a useful property for LLM activations, which vary significantly across layers and inputs. In Section~\ref{sec:midmax} we introduce \emph{Midmax}, a calibration-free alternative to this floor-rounded scale selection.

NVFP4~\citep{cook2025fourover6} uses a two-level scaling strategy. A per-tensor high-precision macroscale $\gamma$ is first computed from the global tensor maximum:

\begin{equation}
    \gamma = \frac{\max(|X|)}{v_{\max}^{\mathrm{FP4}} \cdot v_{\max}^{\mathrm{E4M3}}},
    \label{eq:nvfp4_macro}
\end{equation}

where $v_{\max}^{\mathrm{FP4}}=6$ and $v_{\max}^{\mathrm{E4M3}}=448$ are the maximum representable values of E2M1 and E4M3 respectively. Each block of 16 elements is then assigned an E4M3 block scale by rounding to the nearest representable E4M3 value, using the same $\operatorname{round}_{\mathcal{F}}$ operator from Equation~\ref{eq:quant} with $\mathcal{F} = $ E4M3:
\begin{equation}
    \Delta = \operatorname{round}_{\mathrm{E4M3}}\!\left(\frac{\max_{i=1}^{16} |x_i|}{\gamma \cdot v_{\max}^{\mathrm{FP4}}}\right).
    \label{eq:nvfp4_block}
\end{equation}

E4M3 allows non-power-of-two block scales, providing finer-grained dynamic range coverage within each block.


\subsection{Post-Training Quantization for Large Language Models}\label{sec:background:PTQ}

Post-training quantization (PTQ) has become a key enabler of efficient LLM deployment. GPTQ demonstrated that transformer weights can be quantized to 3-4 bits with minimal accuracy loss~\citep{frantar2022gptq}, and subsequent methods have refined this through alternative reconstruction objectives that better preserve layer outputs~\citep{nagel2020up,li2021brecq}. Complementary approaches address activation quantization: SmoothQuant migrates outlier difficulty from activations to weights via per-channel scaling~\citep{xiao2023smoothquant}, while LLM.int8() isolates outlier channels in higher precision~\citep{dettmers2022gpt3}. Quantization error does not remain localized within individual layers but accumulates and propagates across the network. Sequential correction~\citep{zhang2023post} explicitly accounts for how noise introduced in earlier layers affects downstream computations, and Qronos~\citep{zhang2025qronoscorrectingpastshaping} demonstrates that correcting accumulated prior-layer error through the current quantization decision and updates to yet-to-be-quantized weights can significantly improve low-bit PTQ outcomes.

While these methods have advanced the state-of-the-art in low-bit PTQ, they have largely been developed and evaluated under fixed format configurations. The interaction between PTQ algorithms and BFP configuration choices (block size, scale format, macroscaling, and scale selection) remains underexplored. In this work, PTQ algorithm is one axis of our sweep: we evaluate GPTQ and Qronos alongside format and rotation choices, and compare both against RTN (Round-to-Nearest), a non-learned baseline that independently rounds each value to its nearest representable point without any reconstruction objective.

\subsection{Hadamard Rotations}
\label{sec:hadamard}
Rotation transforms such as the Hadamard transform have become an important preprocessing step for PTQ, redistributing outliers across dimensions to produce smoother, more quantization-friendly distributions~\citep{tseng2024quip, ashkboos2024quarot, liu2025spinquant}. A Hadamard matrix $R$ is an orthogonal matrix with entries $\pm 1/\sqrt{h}$. Each input coordinate is mixed into every output coordinate with a coefficient of magnitude $1/\sqrt{h}$. A large value concentrated in one channel therefore contributes to many rotated coordinates, which lowers the peak channel magnitude that sets the block scale and makes the tensor easier to quantize.

Given an input tensor $X \in \mathbb{R}^{n \times d}$, the rotation is applied as $\tilde{X}=XR$, where $R$ is a Hadamard matrix, and the inverse rotation is absorbed into subsequent weights $W$ as $\tilde{W} = WR$, so the full computation $\tilde{X} \tilde{W}^\top = XW^\top$ is preserved exactly. In practice, the Hadamard matrix dimension $h$ may not be equal to $d$, so $R$ is applied as a block-diagonal matrix with $h \times h$ Hadamard blocks tiling the last dimension of $X$. The choice of block size $h$ determines the granularity of outlier redistribution: smaller blocks (e.g., $h=16$ or $h=32$) rotate within subsets of the feature dimension, while a full-dimensional rotation ($h=d$) redistributes outliers globally, as used in prior work such as QuaRot~\citep{ashkboos2024quarot}. 

\section{PTQ Configurations for Effective MXFP4 Inference}
\label{sec:methods}

We evaluate post-training quantization performance across five configuration axes: block scale format, scale selection,  rotation, PTQ algorithm, and precision. A summary of these configuration settings is in Table~\ref{tab:ptq_configurations}, and we go into more detail below.

\begin{table}[t]
\centering
\caption{Summary of PTQ configuration settings.}
\label{tab:ptq_configurations}
\small
\setlength{\tabcolsep}{10pt}
\renewcommand{\arraystretch}{1.25}
\begin{tabular}{ll}
\toprule

Rotation             & None, Block-16, Block-32, Full \\
PTQ algorithm           & RTN, Qronos, GPTQ \\
Activation precision & 4-bit, 6-bit \\
Weight precision     & 4-bit, 6-bit \\
Block-scale datatype & E8M0, E4M3 \\
Scale selection      & Stats, MSE, Midmax \\
Macroscaling         & Yes, No\\
\bottomrule
\end{tabular}
\end{table}

\paragraph{Macroscaling}
\label{sec:scale_format}
Prior work has studied Macro Block Scaling, which shares a higher-precision scale across $1 \times 128$ elements~\citep{arxiv2603.08713}. The macroscale defined here uses one high-precision scale per tensor. Since the macroscale for weights can be computed offline but the macroscale for activations must be computed online, we use a computationally lighter method for activations and derive $\gamma_{\mathrm{activation}}$ from the tensor-level maximum magnitude:

\begin{equation}
    \gamma_{\mathrm{activation}} = \frac{\max |X|}{v_{\max}^{\mathrm{FP4}}}.
    \label{eq:macro_act}
\end{equation}

To calculate the weight macroscale, $\gamma_{\mathrm{weight}}$, we consider each block's absolute maximum, $m_b$, and $s_b = m_b/v_{\max}^{\mathrm{FP4}}$ as the high-precision scale that can normalize the block values into the FP4 range. An E8M0 block scale has no mantissa bits and thus cannot encode $s_b$ exactly; therefore, we capture this scale with a high-precision macroscale. We calculate $\gamma_{\mathrm{weight}}$ as the geometric mean of all $s_b$ over all $B$ blocks in a tensor. The residual per-block variation relative to $\gamma_{\mathrm{weight}}$ is then captured by the E8M0 block scales within their power-of-two precision:

\begin{equation}
    \gamma_{\mathrm{weight}} = 2^{\,\frac{1}{B}\sum_{b=1}^{B} \log_2\!\left(s_b\right)}.
    \label{eq:macro_w}
\end{equation}

The composite dequantization in both cases is $\hat{x} = \gamma \cdot \Delta_i \cdot x_q$, where $\Delta_i$ is the E8M0 block scale for block $i$ and $\gamma$ is either $\gamma_{\mathrm{activation}}$ or $\gamma_{\mathrm{weight}}$.

\paragraph{Scale Selection}
\label{sec:midmax}

We consider three strategies for choosing the block scale $\Delta$: \texttt{Stats}, a calibration-free strategy that computes the scale directly from the block’s absolute maximum using the floor-rounding rule in Equation~\ref{eq:sharedScale}; \texttt{MSE}, which uses a calibration set to search over a small set of candidate scales and selects the one that minimizes the mean-squared reconstruction error of the dequantized block, following standard calibration-based rounding procedures~\citep{nagel2020up,li2021brecq}; and \textbf{Midmax}, a calibration-free refinement of \texttt{Stats} applicable only to power-of-two scale formats.

For power-of-two block scales, the Stats scale computation applies floor rounding to $E_{\text{block}}$, which forces the maximum absolute value in a block to always be mapped into $[4,8)$ after scaling, so values above $v_{\max}$ are clipped. Any value that falls just above $v_{\max}$ after scaling is hard-clipped, even though it could be accommodated with negligible precision loss by rounding the block scale up to the next power of two. This systematic clipping at the boundary of the representable range has been independently identified in~\citep{arxiv2603.08713}.

We propose Midmax, a calibration-free scale selection strategy that addresses this clipping issue by shifting the rounding decision boundary upward to the midpoint between $v_{\max}$ and the next power of two:
\begin{equation}
    v_{\text{mid}} = \frac{v_{\max} + 2^{e_{\max}+1}}{2}
    \label{eq:vmid}
\end{equation}
where $e_{\max}$ is the largest exponent in the element datatype. When $\max_i |x_i| < v_{\text{mid}}$ (after initial scaling), the tighter scale is retained; when $\max_i |x_i| \geq v_{\text{mid}}$, the scale rounds up to the next power of two. This trades a modest increase in rounding error for a reduction in clipping error on near-boundary values. For E2M1: $v_{\max} = 6.0$ and $v_{\text{mid}} = 7.0$; for E2M3: $v_{\max} = 7.5$ and $v_{\text{mid}} = 7.75$.

Unlike prior work on optimal clipping thresholds for integer quantization~\citep{banner2019post}, Midmax operates within the discrete power-of-two constraint imposed by BFP hardware and requires no changes to the hardware datapath. It applies only to power-of-two scale formats; E4M3 has its own mantissa bits and does not require a boundary correction. For all E8M0 configurations, we apply Midmax to the weights while keeping Stats scale selection for the activations.

\paragraph{Hadamard Rotation}
\label{sec:rotation_configs}

We apply the Hadamard rotation described in Section~\ref{sec:hadamard} at four granularities: no rotation, block-16 ($h=16$), block-32 ($h=32$), and full ($h=d$). We evaluate each granularity across all PTQ algorithms, block-scale formats, and precision configurations in our sweep. Section~\ref{sec:rotation_ptq} shows how the choice of $h$ interacts with both the block-scale format and the PTQ algorithm.

\paragraph{PTQ Algorithms}
\label{sec:ptq_configs}

We evaluate two PTQ algorithms, \texttt{GPTQ} and \texttt{Qronos}, against a non-learned rounding baseline, \texttt{RTN}. \texttt{RTN} (Round-to-Nearest) independently rounds each value to its nearest representable point without any optimization and serves as the baseline against which the two PTQ algorithms are measured. \texttt{GPTQ}~\citep{frantar2022gptq} is a popular error correction method that minimizes layer-wise weight reconstruction error by solving a weighted least-squares problem column by column, compensating for the error incurred in previously quantized columns. \texttt{Qronos}~\citep{zhang2025qronoscorrectingpastshaping} extends this by correcting accumulated prior-layer error through the current quantization decision and updates to yet-to-be-quantized weights, shaping those corrections to account for how quantization noise propagates forward through the network. These methods are evaluated across combinations of precision configuration, scale format, scale selection strategy, and rotation, letting us measure both the main effect of each PTQ algorithm and how each interacts with format and rotation choices.

\paragraph{Precision Configurations}
\label{sec:precision}

We evaluate three precision configurations to isolate the effects of activation and weight precision. In \texttt{A4W4}, both activations and weights use E2M1 (FP4). In \texttt{A6W4}, activations use E2M3 (FP6) while weights remain in E2M1, isolating the effect of increasing activation precision without changing weight storage. In \texttt{A6W6}, both use E2M3, providing a higher-precision reference and measuring the additional benefit of increasing weight precision once activations are in FP6. Together, these configurations trace a two-stage path from \texttt{A4W4} to \texttt{A6W6}, first increasing activation precision and then weight precision.

\begin{figure}
    \centering
    \includegraphics[width=\linewidth]{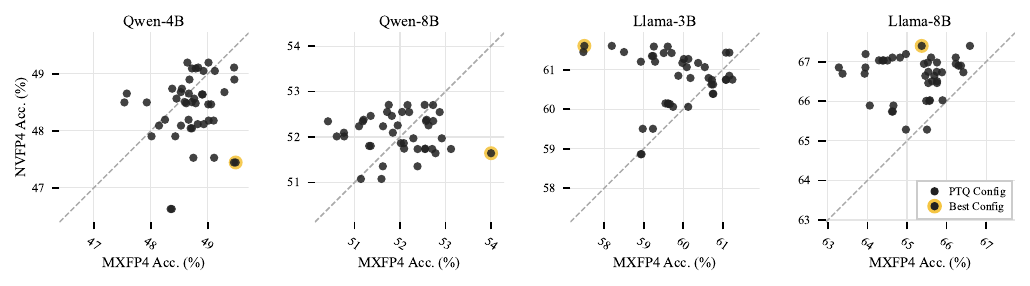}
    \caption{Each point is one PTQ configuration evaluated under MXFP4 ($x$-axis) and NVFP4 ($y$-axis) for a given model; points above the diagonal are configurations where NVFP4 scores higher, and highlighted points mark the best configuration found for each model. The large spread of points within every panel shows that the quantization recipe affects accuracy as much as the choice of block format, and the best MXFP4 and NVFP4 recipes need not coincide. With an appropriate recipe both formats reach competitive accuracy, with neither uniformly ahead across models.}
    \label{fig:mxfp4_vs_nvfp4}
\end{figure}

\section{Results}\label{sec:results}

\paragraph{Experimental Setup}\label{sec:setup}

We evaluate six models spanning two families and three parameter scales: Llama-3.2-1B, Llama-3.2-3B, and Llama-3.1-8B from the Llama family, and Qwen3-1.7B, Qwen3-4B, and Qwen3-8B from the Qwen family. Unless otherwise stated, we use the base model version, and explicitly state whether a model is instruction-tuned. All weights are quantized to the target format and precision while non-linear layers are maintained in BF16. For a given format and precision, we consider the "baseline" PTQ configuration as having no rotation, RTN, MSE scale selection and the default format macroscaling configuration.
Zero-shot accuracy is reported on five commonsense benchmarks: ARC-Easy and ARC-Challenge~\citep{clark2018think}, HellaSwag~\citep{zellers2019hellaswag}, PIQA~\citep{bisk2020piqa}, and WinoGrande~\citep{sakaguchi2021winogrande} and average accuracy across these five tasks as our primary metric. All results are reported using the LM evaluation harness with PyTorch 2.9.1 and ROCm 6.4. For the Llama-3.1-8B instruction-tuned variant, we additionally evaluate on the GSM8K mathematical reasoning benchmark~\citep{cobbe2021gsm8k}. Throughout the paper, all reported accuracy changes (\textit{e.g.}, ``$+1.3\%$'') are absolute differences in percentage points of accuracy, not relative improvements.

\subsection{The Recipe Matters as Much as the Format}
\label{sec:format_comparison}

We begin with the central observation of this study: the accuracy of a microscaling format depends strongly on the quantization recipe. Figure~\ref{fig:mxfp4_vs_nvfp4} shows, for each model, the average benchmark accuracy of every evaluated PTQ configuration for MXFP4 and NVFP4. Two patterns motivate the analysis that follows. First, the accuracy variation across recipes within a format is comparable to or larger than the gap between formats, highlighting the importance of recipe selection. Second, effective recipes are format dependent. For Llama-3.2-3B, for example, the strongest NVFP4 configuration uses Qronos, whereas the strongest MXFP4 configuration uses GPTQ with Midmax scale selection. With suitable recipes, both formats achieve competitive accuracy, and neither consistently leads across models and tasks.

These observations motivate studying how each format responds to the techniques combined with it. The following subsections examine the effects of rotation, PTQ algorithm, scale selection, macroscaling, and precision, including their interactions with the underlying format. Section~\ref{sec:configuration_summary} then summarizes the best-performing recipes for each datatype.

\begin{figure}
    \centering
    \includegraphics[width=\linewidth]{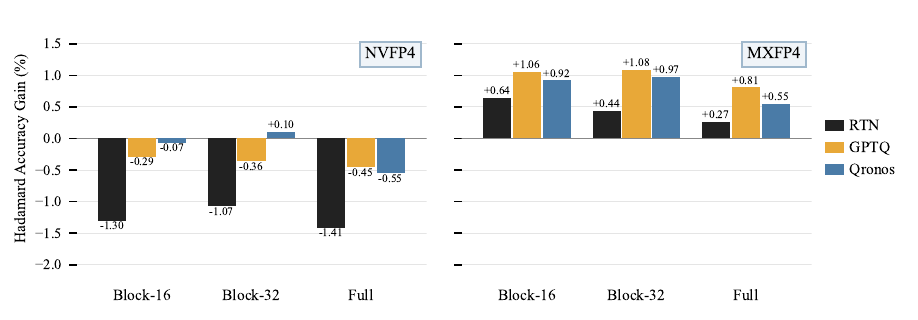}
    \caption{Hadamard rotation $\times$ PTQ method. We consider the average accuracy gain over no rotation for different Hadamard sizes (x-axis) on both NVFP4 (left) and MXFP4 (right) at A4W4. On average, Hadamard rotation does not have a positive effect on NVFP4, yet yields significant benefits for MXFP4. While these represent trends averaged over all  configurations, a best PTQ recipe for a given model may not follow this trend as shown in Table~\ref{tab:scale_config_sensitivity}. }
    \label{fig:rotation_effects}
\end{figure}

\subsection{Rotation and PTQ Algorithm Are Complementary Techniques}
\label{sec:rotation_ptq}

As established above, a recipe tuned for one format does not transfer directly to the other. Hadamard rotation provides a clear example of why. As shown in Figure~\ref{fig:rotation_effects}, rotation affects the two formats very differently at A4W4. For MXFP4, every rotation granularity improves average accuracy across the three PTQ methods, with Block-16 and Block-32 producing the strongest and most consistent gains. When paired with GPTQ or Qronos, these blockwise rotations improve accuracy by approximately $0.9$--$1.1\%$ over no rotation, compared with gains of $0.44$--$0.64\%$ under RTN. This improvement suggests that rotation and PTQ address complementary sources of error: rotation redistributes outliers before quantization, while PTQ reduces the remaining reconstruction error.
NVFP4 shows the opposite trend. Rotation reduces average accuracy in eight of the nine configurations shown in Figure~\ref{fig:rotation_effects}. GPTQ and Qronos moderate this degradation relative to RTN, but do not make rotation consistently beneficial. Block-32 with Qronos is the only configuration that produces a positive average gain, and the improvement is limited to $0.10\%$. This finding is consistent with Egiazarian et al.~\citep{arxiv2509.23202}, who similarly observe that applying Hadamard rotation to NVFP4 degrades performance. One possible explanation is that NVFP4's smaller blocks and finer-grained E4M3 scales already capture more local variation, leaving less residual error for rotation to correct.

The granularity of the rotation further shapes this interaction. Full rotation provides smaller gains than blockwise rotation for MXFP4 and produces some of the largest accuracy reductions for NVFP4. Broad outlier redistribution is therefore not inherently preferable to block-wise rotation. Overall, blockwise rotation is an effective complement to GPTQ and Qronos for MXFP4, while the same preprocessing is largely redundant or detrimental for NVFP4. These aggregate trends provide a useful default, although the model-specific recipes in Table~\ref{tab:scale_config_sensitivity} confirm that no single rotation and PTQ combination is uniformly best.

\begin{figure}[t]
    \centering
    \begin{minipage}[t]{0.64\linewidth}
        \centering
        \includegraphics[width=\linewidth]{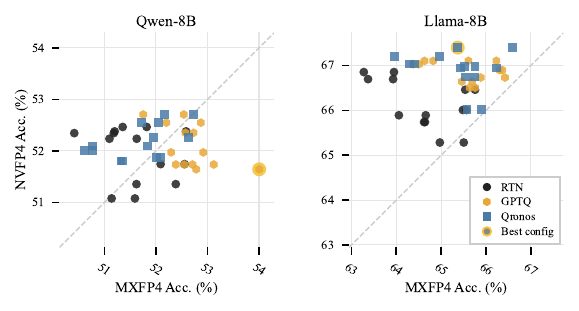}
    \end{minipage}
    \hfill
    \begin{minipage}[t]{0.34\linewidth}
        \centering
        \includegraphics[width=\linewidth]{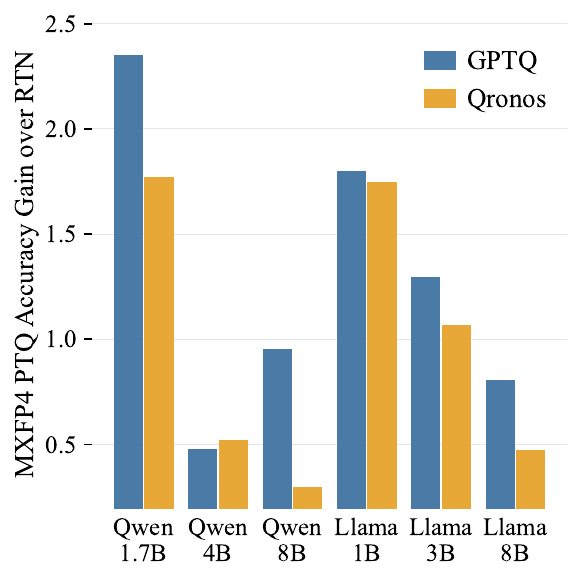}
    \end{minipage}
    \caption{PTQ algorithm performance at A4W4. \textbf{Left and Middle:} The best configuration for Qwen3-8B and Llama-3.1-8B are GPTQ + MXFP4 and Qronos + NVFP4 respectively. Neither PTQ method is  uniformly best by model or datatype. \textbf{Right:} For MXFP4, both GPTQ and Qronos achieve a clear average accuracy gain over RTN, illustrating that PTQ algorithms are an essential tool for maximizing quantized model performance.}
    \label{fig:ptq_comparison}
\end{figure}

\subsection{PTQ Algorithm Effectiveness Depends on the Model and Format}
\label{sec:ptq_results}

We next isolate the effect of the PTQ algorithm itself. Figure~\ref{fig:ptq_comparison} shows that neither GPTQ nor Qronos is uniformly optimal across models or formats. For Qwen3-8B, the best overall configuration combines GPTQ with MXFP4, whereas Llama-3.1-8B achieves its highest accuracy using Qronos with NVFP4. The distribution of configurations further shows that a PTQ method that performs well for one format does not necessarily transfer to the other. Despite this configuration dependence, both methods provide clear gains over RTN. Averaged across the full MXFP4 sweep, GPTQ improves accuracy by $1.32\%$, while Qronos provides a $1.03\%$ gain. NVFP4 also benefits, although more modestly, with average gains of $0.78\%$ for GPTQ and $0.88\%$ for Qronos. PTQ therefore provides greater average accuracy recovery for MXFP4, particularly with GPTQ, indicating that its coarser power-of-two scaling leaves more quantization error that can be corrected through reconstruction-aware optimization. Furthermore, Figure~\ref{fig:ptq_comparison} (right) shows that the magnitude of the PTQ gain varies substantially across models. Smaller models generally show greater accuracy improvement from GPTQ and Qronos, while the gains narrow at larger model scales. Nevertheless, both techniques remain beneficial across the evaluated MXFP4 models.

\begin{figure}
    \centering
    \includegraphics[width=\linewidth]{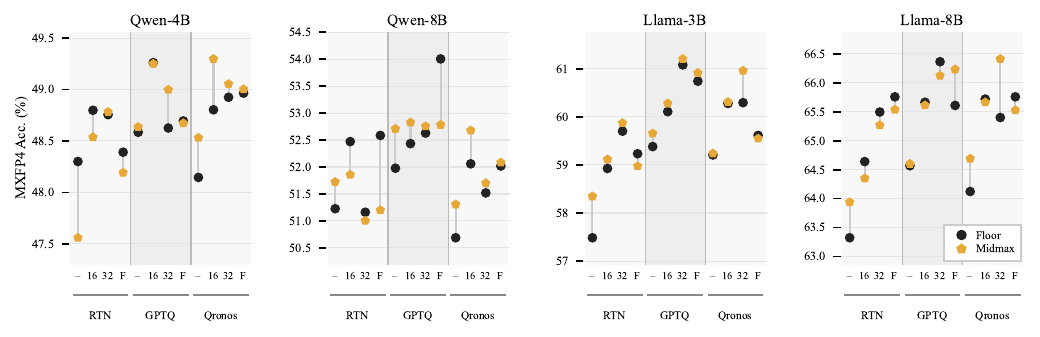}
    \caption{Midmax performance across different choices of Hadamard size \{None (-), 16, 32, Full (F)\} and PTQ methods \{RTN, GPTQ, Qronos\} for Qwen and Llama model families. Midmax has the capacity to give a significant boost in average accuracy for MXFP4 at A4W4, particularly for configurations with Qronos. }
    \label{fig:midmax_results}
\end{figure}

\subsection{Midmax Provides Calibration-Free Scale Selection}
\label{sec:midmax_results}

Figure~\ref{fig:midmax_results} compares Midmax with standard floor-based scale selection across models, rotation granularities, and PTQ methods. Averaged over the full sweep, Midmax improves accuracy by $0.12\%$ and outperforms floor rounding in $67\%$ of the evaluated model, rotation, and PTQ combinations. Although the average improvement is modest, the consistency of these gains shows that scale selection remains relevant even after rotation and PTQ are applied. Furthermore, the benefit of Midmax depends strongly on the surrounding configuration. Without rotation, Midmax provides an average gain of $0.25\%$, including an improvement of $0.62\%$ on Qwen3-8B. The gain becomes negligible with Block-16 rotation but increases again to $0.20\%$ with Block-32 rotation. Midmax and Hadamard rotation are therefore neither fully additive nor fully redundant. Both reduce errors caused by outliers, but they act at different stages of quantization and remain complementary in selected configurations.

The interaction with PTQ is even more pronounced. Midmax improves accuracy by $0.30\%$ when paired with Qronos, compared with $0.06\%$ for GPTQ and $0.02\%$ for RTN. This suggests that reducing clipping near power-of-two scale boundaries remains valuable after Qronos corrects reconstruction and error-propagation effects. Midmax is not uniformly superior to calibrated strategies such as MSE, but it provides a reliable improvement over floor rounding without calibration data or additional preprocessing. It is therefore most useful as a low-cost scale-selection strategy for sensitive MXFP4 configurations, particularly when residual outliers remain after rotation.

\begin{table}[t]
    \centering
    \small
    \setlength{\tabcolsep}{9pt}
    \renewcommand{\arraystretch}{1.2}
    \begin{tabular}{llccc}
        \toprule
        Format & Source & WinoGrande & HellaSwag & GSM8K \\
        \midrule
        BF16                  & in-house  & 78.1 & 80.1 & 85.4 \\
        \midrule
        NVFP4                 & Egiazarian et al.~\citep{arxiv2509.23202} & 75.2 & 79.5 & 82.6 \\
        NVFP4                 & in-house  & 75.6 & 79.3 & 82.7 \\
        \midrule
        MXFP4                 & Egiazarian et al.~\citep{arxiv2509.23202} & 74.8 & 75.3 & 75.7 \\
        MXFP4                 & in-house  & 77.0 & 78.7 & 79.6 \\
        MXFP4 + macroscale    & in-house  & 77.2 & 78.8 & 81.9 \\
        \midrule
        MXFP6 act + MXFP4 wt & in-house  & \textbf{78.3} & \textbf{80.1} & \textbf{85.7} \\
        \bottomrule
    \end{tabular}
    \caption{Accuracy (\%) on WinoGrande, HellaSwag, and GSM8K for Llama-3.1-8B instruction-tuned. ``in-house'' uses our best configurations. Each Egiazarian et al.~\citep{arxiv2509.23202} entry is the best result reported for that task and format and may come from a different PTQ method.}
    \label{tab:instruct}
\end{table}

\begin{figure}[t]
    \centering
    \includegraphics[width=\linewidth]{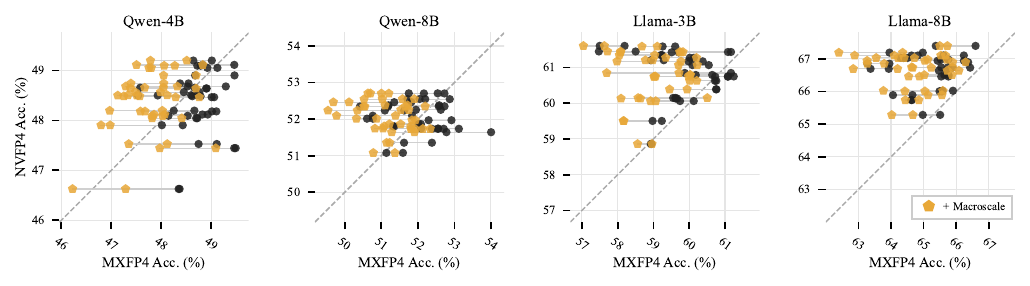}
    \caption{\textbf{Effect of NVFP4-style macroscaling on MXFP4.} Under the same experimental settings as in Figure~\ref{fig:mxfp4_vs_nvfp4}, we compare the original MXFP4 results with MXFP4 augmented by NVFP4-style macroscaling. For each configuration, the horizontal shift represents the resulting change in MXFP4 accuracy due to applying macroscale. Among the best-performing MXFP4 PTQ configurations, macroscaling does not improve average accuracy.}
    \label{fig:macroscale}
\end{figure}

\subsection{Applying NVFP4-Style Macroscaling to MXFP4}
\label{sec:macroscale_results}

We augment MXFP4 with a software-assisted per-tensor macroscale and compare the resulting MXFP4+macroscale configuration with standard MXFP4. This modification does not convert MXFP4 into NVFP4, since the two formats retain different block sizes and block-scale datatypes, but instead allows us to test whether a scaling mechanism that is intrinsic to NVFP4 also benefits MXFP4.
At A4W4, macroscaling generally reduces average benchmark accuracy for MXFP4. Figure~\ref{fig:macroscale} compares MXFP4 configurations with and without macroscaling across Qwen3-4B, Qwen3-8B, Llama-3.2-3B, and Llama-3.1-8B. Across all models, the macroscaled configurations lie predominantly to the left of their standard MXFP4 counterparts, corresponding to mean reductions of $0.66$-$0.85\%$, depending on the model. This trend also holds among the highest-accuracy MXFP4 configurations: macroscaling does not systematically move them toward parity with NVFP4 and often widens the gap.

The effect is not entirely uniform across configurations, and a small number of rightward shifts in Figure~\ref{fig:macroscale} indicate isolated improvements from macroscaling, but these gains are insufficient to reverse the aggregate trend for any of the four models. Thus, although macroscaling is an essential component of the complete NVFP4 format, its benefits do not transfer directly to MXFP4. 
Despite the decline in average accuracy, macroscaling can benefit individual tasks. On Llama-3.1-8B-Instruct, it raises our best MXFP4 accuracy from $77.0\%$ to $77.2\%$ on WinoGrande, from $78.7\%$ to $78.8\%$ on HellaSwag, and from $79.6\%$ to $81.9\%$ on GSM8K (Table~\ref{tab:instruct}). The largest gain is therefore on GSM8K, with smaller improvements on the two commonsense benchmarks. For context, the macroscaled MXFP4 results exceed the best task-wise NVFP4 result reported by Egiazarian et al.~\citep{arxiv2509.23202} on WinoGrande ($77.2\%$ vs.\ $75.2\%$), while remaining below their results on HellaSwag ($78.8\%$ vs.\ $79.5\%$) and GSM8K ($81.9\%$ vs.\ $82.6\%$). These comparisons use the best reported result for each task and format, so they do not necessarily reflect a single PTQ recipe.

\begin{table}[t]
    \centering
    \small
    \setlength{\tabcolsep}{10pt}
    \renewcommand{\arraystretch}{1.08}
    \caption{\textbf{General trends observed across PTQ configurations.} We report change in average accuracy for the principal configuration
    choices evaluated in this study. Results are averaged over the remaining compatible configurations.}
    \label{tab:results_summary}
    \begin{tabular}{@{}lcc@{}}
        \toprule
        Configuration change & MXFP4 & NVFP4 \\
        \midrule

        Activation precision: A4W4 to A6W4
            & $+1.74$ & -- \\
        Weight precision: A6W4 to A6W6
            & $+1.26$ & -- \\

        \addlinespace[2pt]
        GPTQ over RTN
            & $+1.32$ & $+0.78$ \\
        Block 32 over no rotation
            & $+1.17$ & $-0.45$ \\
        Midmax over floor, without rotation
            & $+0.25$ & -- \\
        Adding a macroscale at A4W4
            & $-0.66$ & \textit{built in} \\

        \bottomrule
    \end{tabular}
\end{table}

\subsection{Configuration Choices Must Be Evaluated Jointly}
\label{sec:configuration_summary}

Table~\ref{tab:results_summary} places the principal effects observed in our configuration sweep on a common scale. Among changes that retain A4W4 precision, GPTQ and Block-32 rotation provide the largest average gains for MXFP4, improving accuracy by $1.32\%$ and $1.17\%$, respectively. Midmax provides a smaller but calibration-free improvement of $0.25\%$ without rotation. The corresponding effects differ for NVFP4: GPTQ provides a smaller gain of $0.78\%$, while Block-32 rotation reduces average accuracy by $0.45\%$. These differences explain why the same quantization recipe does not transfer uniformly between the two formats.

When additional precision is available, increasing activation precision from A4W4 to A6W4 improves MXFP4 accuracy by $1.74\%$, and subsequently increasing weight precision from A6W4 to A6W6 provides an additional $1.26\%$ gain.

\begin{table*}
    \centering
    \small
    \setlength{\tabcolsep}{9pt}
    \renewcommand{\arraystretch}{1.2}
    \caption{Best-performing configurations for MXFP4 and NVFP4 across models.}
    \label{tab:best_format_configs}
    \begin{tabular}{@{}l lll @{\hspace{12pt}} lll@{}}
        \toprule
        \multirow{2}{*}{Model}
        & \multicolumn{3}{c}{MXFP4}
        & \multicolumn{3}{c}{NVFP4} \\
        \cmidrule(lr){2-4}
        \cmidrule(lr){5-7}
        & PTQ Method & Rotation & Scale
        & PTQ Method & Rotation & Scale \\
        \midrule

        Qwen3-1.7B
        & GPTQ   & Block 16 & MSE
        & Qronos & Block 32 & Stats \\

        Qwen3-4B
        & GPTQ   & Block 16 & Midmax
        & Qronos & None     & Stats \\

        Qwen3-8B
        & GPTQ   & Full     & MSE
        & Qronos & Full     & Stats \\

        \midrule

        Llama-3.2-1B
        & GPTQ   & Block 16 & MSE
        & GPTQ   & Block 16 & MSE \\

        Llama-3.2-3B
        & GPTQ   & Block 32 & Midmax
        & Qronos & None     & Stats \\

        Llama-3.1-8B
        & Qronos & Block 32 & Midmax
        & Qronos & Block 32 & MSE \\

        \bottomrule
    \end{tabular}
\end{table*}

The average effects provide useful guidance, but they do not define a universal recipe. Table~\ref{tab:best_format_configs} shows that GPTQ is selected by the best MXFP4 configuration for five of the six models, whereas Qronos is selected by the best NVFP4 configuration for five of the six models. Blockwise rotation appears in five of the six best MXFP4 configurations, while the preferred NVFP4 rotation spans none, Block-16, Block-32, and full rotation. Scale selection is similarly model dependent. The best MXFP4 configurations are evenly divided between MSE and Midmax, while NVFP4 primarily favors Stats but selects MSE for two Llama models.

These results point to a useful default recipe, though, as Table~\ref{tab:best_format_configs} shows, not a single configuration that is optimal for every model. For MXFP4 at A4W4, PTQ and blockwise rotation provide the largest and most consistent gains, with Midmax offering an additional calibration-free improvement. Their combined contribution raises accuracy by approximately $3\%$ over baseline. For NVFP4, Qronos is selected more frequently, while rotation requires greater model-specific care. When this A4W4 recipe does not provide sufficient accuracy, moving to A6W4 is an effective next step, though we did not evaluate the complementary A4W6 configuration needed to determine whether increasing weight precision first would be equally or more effective. Macroscaling should instead be reserved for the tasks and precision regimes where its targeted benefits justify adding it to MXFP4.
The appropriate unit of comparison is therefore the complete quantization configuration rather than the block format alone. Format, PTQ algorithm, rotation granularity, scale selection, macroscaling, and element precision jointly determine the final accuracy.

\section{Conclusion}
Accurate microscaling quantization requires jointly choosing the numerical format, PTQ algorithm, rotation, and scale-selection strategy. Across the six language models evaluated, neither MXFP4 nor NVFP4 consistently achieves higher accuracy, and their strongest recipes differ.
Our results offer three practical takeaways. First, GPTQ with blockwise rotation provides a strong starting point for MXFP4, while NVFP4 more often favors Qronos and generally benefits less from rotation. Second, calibration-free Midmax improves upon floor-based scale selection in \(67\%\) of the evaluated configurations, whereas adding a per-tensor macroscale to MXFP4 does not improve average accuracy despite selective task-level gains. Third, A6W4 recovers accuracy while retaining 4-bit weight storage, matching or exceeding our BF16 baseline on the three evaluated tasks for Llama-3.1-8B-Instruct.
Together, these findings support selecting and evaluating complete quantization recipes, using format-specific defaults as starting points for tuning to each model.

\newpage

\bibliographystyle{ieeetr}
\bibliography{references}

@misc{cook2025fourover6,
  title={Four Over Six: More Accurate {NVFP4} Quantization with Adaptive Block Scaling},
  author={Cook, Jack and Guo, Junxian and Xiao, Guangxuan and Lin, Yujun and Wyss, Keith and Nazemi, Mahdi and Mishra, Asit and del Mundo, Carlo and Blankevoort, Tijmen and Han, Song},
  year={2025},
  eprint={2512.02010},
  archivePrefix={arXiv},
  url={https://arxiv.org/abs/2512.02010},
}

@misc{rouhani2023microscaling,
  title={Microscaling Data Formats for Deep Learning},
  author={Rouhani, Bita Darvish and Zhao, Ritchie and More, Ankit and Hall, Mathew and Khodamoradi, Alireza and Deng, Summer and Choudhary, Dhruv and Cornea, Marius and Dellinger, Eric and Denolf, Kristof and others},
  year={2023},
  eprint={2310.10537},
  archivePrefix={arXiv},
  primaryClass={cs.LG},
  url={https://arxiv.org/abs/2310.10537},
}

@inproceedings{arxiv2509.23202,
  title={Bridging the Gap Between Promise and Performance for Microscaling {FP4} Quantization},
  author={Egiazarian, Vage and Castro, Roberto L. and Kuznedelev, Denis and Panferov, Andrei and Kurtic, Eldar and Pandit, Shubhra and Marques, Alexandre and Kurtz, Mark and Ashkboos, Saleh and Hoefler, Torsten and Alistarh, Dan},
  booktitle={International Conference on Learning Representations},
  year={2026},
  eprint={2509.23202},
  archivePrefix={arXiv},
  url={https://arxiv.org/abs/2509.23202},
}

@article{cobbe2021gsm8k,
  title={Training Verifiers to Solve Math Word Problems},
  author={Cobbe, Karl and Kosaraju, Vineet and Bavarian, Mohammad and Chen, Mark and Jun, Heewoo and Kaiser, Lukasz and Plappert, Matthias and Tworek, Jerry and Hilton, Jacob and Nakano, Reiichiro and Hesse, Christopher and Schulman, John},
  journal={arXiv preprint arXiv:2110.14168},
  year={2021}
}

@manual{amd-cdna4-isa,
  title        = {{AMD} Instinct {CDNA4} Instruction Set Architecture Reference Guide},
  organization = {Advanced Micro Devices, Inc.},
  year         = {2025},
  url          = {https://www.amd.com/content/dam/amd/en/documents/instinct-tech-docs/instruction-set-architectures/amd-instinct-cdna4-instruction-set-architecture.pdf},
}

@inproceedings{arxiv2603.08713,
  title={Unveiling the Potential of Quantization with {MXFP4}: Strategies for Quantization Error Reduction},
  author={Chhugani, Jatin and Jeong, Geonhwa and Su, Bor-Yiing and Pan, Yunjie and Yang, Hanmei and Ankit, Aayush and Yu, Jiecao and Deng, Summer and Chen, Yunqing and Satish, Nadathur and Kim, Changkyu},
  booktitle={International Conference on Machine Learning},
  year={2026},
  eprint={2603.08713},
  archivePrefix={arXiv},
  url={https://arxiv.org/abs/2603.08713},
}

@inproceedings{zhang2025qronoscorrectingpastshaping,
      title={Qronos: Correcting the Past by Shaping the Future... in Post-Training Quantization},
      author={Shihao Zhang and Haoyu Zhang and Ian Colbert and Rayan Saab},
      booktitle={International Conference on Learning Representations},
      year={2026},
      eprint={2505.11695},
      archivePrefix={arXiv},
      primaryClass={cs.LG},
      url={https://arxiv.org/abs/2505.11695}, 
}

@article{micikevicius2022fp8,
  title={FP8 formats for deep learning},
  author={Micikevicius, Paulius and Stosic, Dusan and Burgess, Neil and Cornea, Marius and Dubey, Pradeep and Grisenthwaite, Richard and Ha, Sangwon and Heinecke, Alexander and Judd, Patrick and Kamalu, John and others},
  journal={arXiv preprint arXiv:2209.05433},
  year={2022}
}

@inproceedings{bisk2020piqa,
  title={{PIQA}: Reasoning about Physical Commonsense in Natural Language},
  author={Bisk, Yonatan and Zellers, Rowan and Le Bras, Ronan and Gao, Jianfeng and Choi, Yejin},
  booktitle={Proceedings of the AAAI Conference on Artificial Intelligence},
  pages={7432--7439},
  year={2020}
}

@article{sakaguchi2021winogrande,
  title={WinoGrande: An adversarial winograd schema challenge at scale},
  author={Sakaguchi, Keisuke and Bras, Ronan Le and Bhagavatula, Chandra and Choi, Yejin},
  journal={Communications of the ACM},
  volume={64},
  number={9},
  pages={99--106},
  year={2021},
  publisher={ACM New York, NY, USA}
}

@article{clark2018think,
  title={Think you have solved question answering? try ARC, the AI2 reasoning challenge},
  author={Clark, Peter and Cowhey, Isaac and Etzioni, Oren and Khot, Tushar and Sabharwal, Ashish and Schoenick, Carissa and Tafjord, Oyvind},
  journal={arXiv preprint arXiv:1803.05457},
  year={2018}
}

@inproceedings{zellers2019hellaswag,
  title={Hellaswag: Can a machine really finish your sentence?},
  author={Zellers, Rowan and Holtzman, Ari and Bisk, Yonatan and Farhadi, Ali and Choi, Yejin},
  booktitle={Proceedings of the 57th annual meeting of the association for computational linguistics},
  pages={4791--4800},
  year={2019}
}

@inproceedings{banner2019post,
  title={Post Training 4-Bit Quantization of Convolutional Networks for Rapid-Deployment},
  author={Banner, Ron and Nahshan, Yury and Soudry, Daniel},
  booktitle={Advances in Neural Information Processing Systems},
  volume={32},
  year={2019}
}

@inproceedings{xiao2023smoothquant,
  title={SmoothQuant: Accurate and efficient post-training quantization for large language models},
  author={Xiao, Guangxuan and Lin, Ji and Seznec, Mickael and Wu, Hao and Demouth, Julien and Han, Song},
  booktitle={International conference on machine learning},
  pages={38087--38099},
  year={2023},
  organization={PMLR}
}

@article{frantar2022gptq,
  title={{GPTQ}: Accurate Post-training Compression for Generative Pretrained Transformers}, 
  author={Elias Frantar and Saleh Ashkboos and Torsten Hoefler and Dan Alistarh},
  year={2022},
  journal={arXiv preprint arXiv:2210.17323}
}

@inproceedings{tseng2024quip,
  title={Quip\#: Even better llm quantization with hadamard incoherence and lattice codebooks},
  author={Tseng, Albert and Chee, Jerry and Sun, Qingyao and Kuleshov, Volodymyr and De Sa, Christopher},
  journal={Proceedings of machine learning research},
  volume={235},
  pages={48630},
  year={2024}
}

@inproceedings{liu2025spinquant,
  title={Spinquant: Llm quantization with learned rotations},
  author={Liu, Zechun and Zhao, Changsheng and Fedorov, Igor and Soran, Bilge and Choudhary, Dhruv and Krishnamoorthi, Raghuraman and Chandra, Vikas and Tian, Yuandong and Blankevoort, Tijmen},
  booktitle={International Conference on Learning Representations},
  year={2025}
}

@article{ashkboos2024quarot,
  title={Quarot: Outlier-free 4-bit inference in rotated llms},
  author={Ashkboos, Saleh and Mohtashami, Amirkeivan and Croci, Maximilian L and Li, Bo and Cameron, Pashmina and Jaggi, Martin and Alistarh, Dan and Hoefler, Torsten and Hensman, James},
  journal={Advances in Neural Information Processing Systems},
  volume={37},
  pages={100213--100240},
  year={2024}
}

@article{dettmers2022gpt3,
  title={{LLM}.int8(): 8-bit Matrix Multiplication for Transformers at Scale},
  author={Dettmers, Tim and Lewis, Mike and Belkada, Younes and Zettlemoyer, Luke},
  journal={Advances in neural information processing systems},
  volume={35},
  pages={30318--30332},
  year={2022}
}

@inproceedings{li2021brecq,
  title={{BRECQ}: Pushing the Limit of Post-Training Quantization by Block Reconstruction},
  author={Li, Yuhang and Gong, Ruihao and Tan, Xu and Yang, Yang and Hu, Peng and Zhang, Qi and Yu, Fengwei and Wang, Wei and Gu, Shi},
  booktitle={International Conference on Learning Representations},
  year={2021},
  eprint={2102.05426},
  archivePrefix={arXiv},
  url={https://arxiv.org/abs/2102.05426},
}

@inproceedings{nagel2020up,
  title={Up or down? adaptive rounding for post-training quantization},
  author={Nagel, Markus and Amjad, Rana Ali and Van Baalen, Mart and Louizos, Christos and Blankevoort, Tijmen},
  booktitle={International conference on machine learning},
  pages={7197--7206},
  year={2020},
  organization={PMLR}
}

@article{zhang2023post,
  title={Post-training quantization for neural networks with provable guarantees},
  author={Zhang, Jinjie and Zhou, Yixuan and Saab, Rayan},
  journal={SIAM journal on mathematics of data science},
  volume={5},
  number={2},
  pages={373--399},
  year={2023},
  publisher={SIAM}
}

@article{dao2022flashattention,
  title={Flashattention: Fast and memory-efficient exact attention with io-awareness},
  author={Dao, Tri and Fu, Dan and Ermon, Stefano and Rudra, Atri and R{\'e}, Christopher},
  journal={Advances in neural information processing systems},
  volume={35},
  pages={16344--16359},
  year={2022}
}

@article{zheng2024sglang,
  title={SGLang: Efficient execution of structured language model programs},
  author={Zheng, Lianmin and Yin, Liangsheng and Xie, Zhiqiang and Sun, Chuyue Livia and Huang, Jeff and Yu, Cody Hao and Cao, Shiyi and Kozyrakis, Christos and Stoica, Ion and Gonzalez, Joseph E and others},
  journal={Advances in neural information processing systems},
  volume={37},
  pages={62557--62583},
  year={2024}
}

@misc{OCP-MX-v1.0,
  title        = {OCP Microscaling Formats (MX) Specification Version 1.0},
  author       = {Darvish Rouhani, Bita and Garegrat, Nitin and Savell, Tom and More, Ankit and Han, Kyung-Nam and Zhao, Ritchie and Hall, Mathew and Klar, Jasmine and Chung, Eric and Yu, Yuan and Schulte, Michael and Wittig, Ralph and Bratt, Ian and Stephens, Nigel and Milanovic, Jelena and Brothers, John and Dubey, Pradeep and Cornea, Marius and Rodriguez, Andres and Langhammer, Martin and Deng, Summer and Naumov, Maximilian and Micikevicius, Paulius and Siu, Michael and Verrilli, Colin and others},
  year         = {2023},
  howpublished = {\url{https://www.opencompute.org/documents/ocp-microscaling-formats-mx-v1-0-spec-final-pdf}},
  note         = {Open Compute Project Specification, accessed 2025-01-19}
}

@article{stojkovic2024towards,
  title={Towards greener llms: Bringing energy-efficiency to the forefront of llm inference},
  author={Stojkovic, Jovan and Choukse, Esha and Zhang, Chaojie and Goiri, Inigo and Torrellas, Josep},
  journal={arXiv preprint arXiv:2403.20306},
  year={2024}
}

@inproceedings{firoozshahian2023mtia,
  title={Mtia: First generation silicon targeting meta's recommendation systems},
  author={Firoozshahian, Amin and Coburn, Joel and Levenstein, Roman and Nattoji, Rakesh and Kamath, Ashwin and Wu, Olivia and Grewal, Gurdeepak and Aepala, Harish and Jakka, Bhasker and Dreyer, Bob and others},
  booktitle={Proceedings of the 50th Annual International Symposium on Computer Architecture},
  pages={1--13},
  year={2023}
}
\newpage

\appendix

\section{Average Accuracy Across All PTQ Configurations}

\begin{figure}[h]
    \centering
    \includegraphics[width=\linewidth]{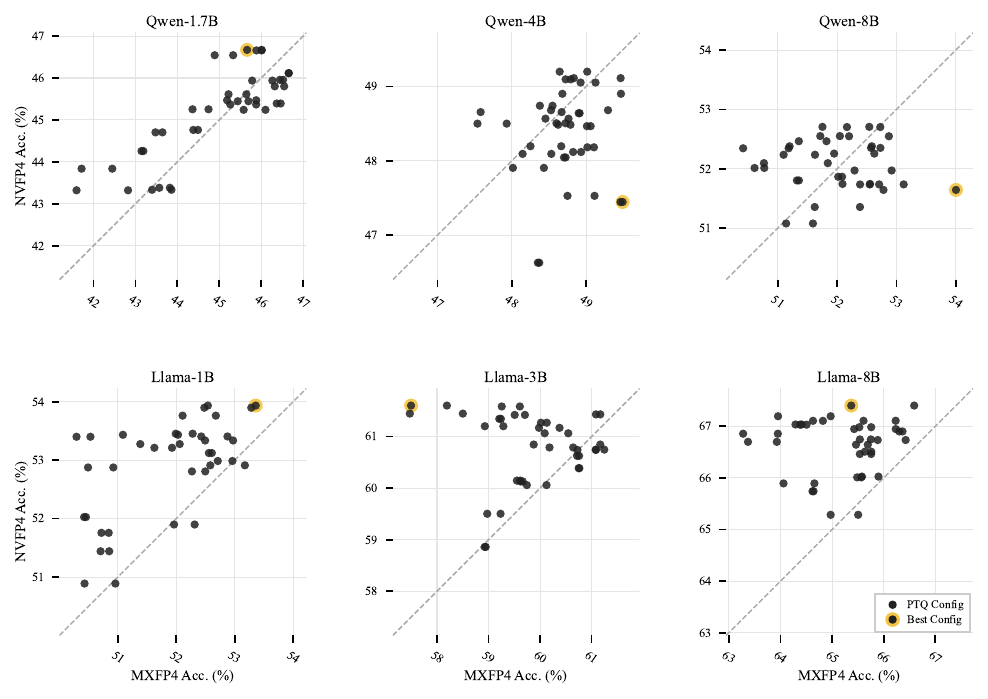}
    \caption{Average benchmark accuracy across PTQ configurations for MXFP4 ($x$-axis) and NVFP4 ($y$-axis), extended to all six models; points above the diagonal are configurations where NVFP4 scores higher, and highlighted points mark the best configuration found for each model. MXFP4 reaches the highest peak accuracy on Qwen3-4B and Qwen3-8B, while NVFP4 reaches the highest peak on Qwen3-1.7B and the evaluated Llama models. The broad spread within each panel shows that the PTQ recipe affects accuracy as much as the choice of block format.}
    \label{fig:mxfp4_vs_nvfp4_all}
\end{figure}

We extend the comparison in Figure~\ref{fig:mxfp4_vs_nvfp4} to additional models from the Qwen3 and Llama families. Figure~\ref{fig:mxfp4_vs_nvfp4_all} presents a paired comparison in which each point represents the same model and PTQ configuration evaluated using MXFP4 and NVFP4. The distance from the diagonal therefore measures the effect of changing the block format while holding the remaining quantization choices fixed.

The expanded results reinforce that neither format is uniformly superior. MXFP4 reaches the highest peak accuracy on Qwen3-4B and Qwen3-8B, while NVFP4 reaches the highest peak on Qwen3-1.7B and the evaluated Llama models. At the same time, the broad spread within each model shows that PTQ configuration can affect accuracy as much as the choice between MXFP4 and NVFP4. Consequently, comparisons based on a single shared PTQ recipe may not reflect the best performance achievable by either format.

\section{NVFP4-style Macroscale Performance on MXFP4}

A central difference between NVFP4 and MXFP4 is how the two formats represent scale. NVFP4 combines E4M3 block scales with a high-precision per-tensor macroscale~\citep{arxiv2509.23202}, whereas standard MXFP4 uses E8M0 block scales without an additional tensor-level scale. This raises a natural question: does the macroscale benefit come specifically from its interaction with the NVFP4 format, or can it also improve MXFP4? To isolate this design choice, we augment MXFP4 with an analogous software-assisted macroscale and compare it with otherwise identical standard MXFP4 configurations. The resulting format remains distinct from NVFP4 because its block size and block-scale datatype are unchanged.

\begin{figure}[h]
    \centering
    \includegraphics[width=\linewidth]{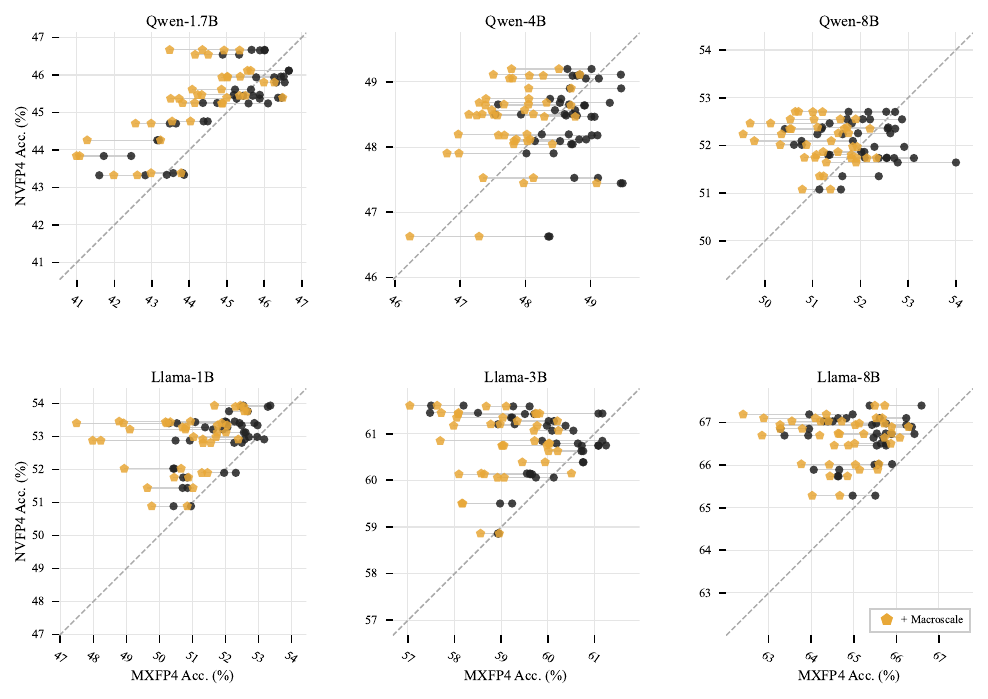}
    \caption{\textbf{Effect of NVFP4-style macroscaling on MXFP4: Average Accuracy.} Using the same PTQ configurations as in Figure~\ref{fig:mxfp4_vs_nvfp4_all}, we compare the original MXFP4 results with those obtained after applying macroscaling on each model. Horizontal shifts indicate the change in MXFP4 accuracy due to macroscale for a given configuration. Results are shown for average accuracy. Although NVFP4-style macroscaling generally reduces average accuracy for MXFP4, it provides task-dependent gains, particularly on WinoGrande (see Figure~\ref{fig:macroscale_wino_all}), demonstrating that its effectiveness depends on both the PTQ configuration and downstream task.}
    \label{fig:macroscale_avg_all}
\end{figure}

The aggregate results initially suggest that the benefit does not transfer. Figure~\ref{fig:macroscale_avg_all} shows that, at A4W4, adding a macroscale generally reduces average benchmark accuracy. For Llama-3.2-1B, the broader configuration sweep produces average reductions of $0.66\%$ to $0.85\%$. Although macroscaling is an integral part of NVFP4, it therefore does not act as a general improvement when added to MXFP4. Its interaction with MXFP4's larger blocks and power-of-two E8M0 scales appears to differ from its role within NVFP4.

\begin{figure}[t]
    \centering
    \includegraphics[width=\linewidth]{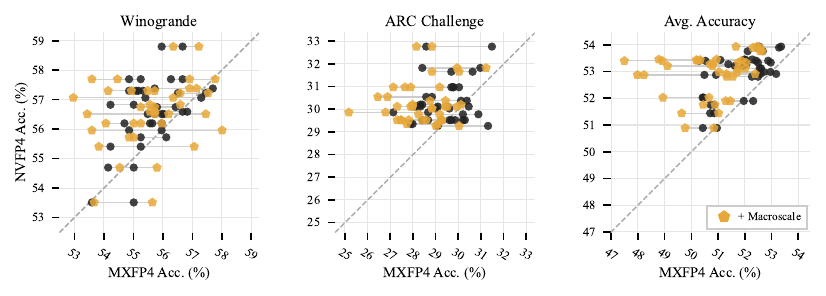}
    \caption{\textbf{Effect of NVFP4-style macroscaling on MXFP4.} Using the same PTQ configurations as in Figure~\ref{fig:mxfp4_vs_nvfp4}, we compare the original MXFP4 results with those obtained after applying macroscaling on Llama-3.2-1B. Horizontal shifts indicate the change in MXFP4 accuracy due to macroscale for a given configuration. Results are shown for WinoGrande, ARC-Challenge, and average accuracy. Although NVFP4-style macroscaling generally reduces average accuracy for MXFP4, it provides task-dependent gains, particularly on WinoGrande, demonstrating that its effectiveness depends on both the PTQ configuration and downstream task.}
    \label{fig:macroscale_per_task}
\end{figure}

\begin{figure}[h]
    \centering
    \includegraphics[width=\linewidth]{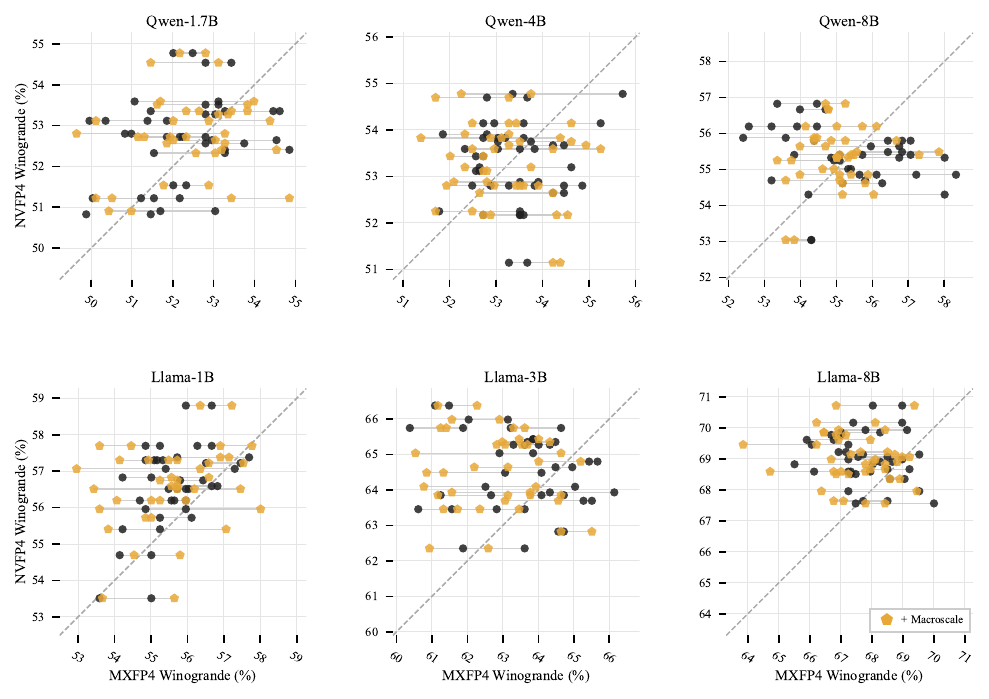}
    \caption{\textbf{Effect of NVFP4-style macroscaling on MXFP4: WinoGrande.} Same as Figure~\ref{fig:macroscale_avg_all} but evaluated on WinoGrande task.}
    \label{fig:macroscale_wino_all}
\end{figure}

The average, however, hides a more selective task-level effect. On WinoGrande as shown in Figures~\ref{fig:macroscale_per_task} and~\ref{fig:macroscale_wino_all}, macroscaling improves Llama-3.2-1B accuracy across every evaluated configuration, even when aggregate accuracy declines. The shifts on ARC-Challenge and average accuracy are smaller and less consistent. Thus, macroscaling does not improve MXFP4 uniformly; instead, it changes the quantization error in ways that can benefit particular downstream tasks.

This conditional benefit also appears across precision regimes and model settings. At A6W6, the average effect reverses, with macroscaling providing a mean improvement of $0.69\%$. Table~\ref{tab:instruct} provides a further example on instruction-tuned Llama-3.1-8B and includes the best task-wise MXFP4 and NVFP4 results reported by Egiazarian et al.~\citep{arxiv2509.23202}; the published entries may use different PTQ methods across tasks. Our best standard MXFP4 scores are $77.0\%$, $78.7\%$, and $79.6\%$ on WinoGrande, HellaSwag, and GSM8K, respectively, while our best macroscaled MXFP4 scores are $77.2\%$, $78.8\%$, and $81.9\%$. Relative to the best reported NVFP4 result for each task, MXFP4 with macroscaling is higher on WinoGrande ($77.2\%$ vs.\ $75.2\%$) and lower on HellaSwag ($78.8\%$ vs.\ $79.5\%$) and GSM8K ($81.9\%$ vs.\ $82.6\%$). Our in-house NVFP4 configuration is also higher than MXFP4 with macroscaling on HellaSwag and GSM8K.

These results position NVFP4-style macroscaling as a targeted extension to MXFP4 rather than a universally beneficial addition. Its value depends on the downstream task and precision regime; its clearest gain in our evaluation is on GSM8K for Llama-3.1-8B-Instruct, where tensor-level range variation may remain important after block scaling, but we have not evaluated enough models and math benchmarks to claim this pattern holds for mathematical reasoning in general.

\section{PTQ Method}

\begin{figure}[h]
    \centering
    \includegraphics[width=\linewidth]{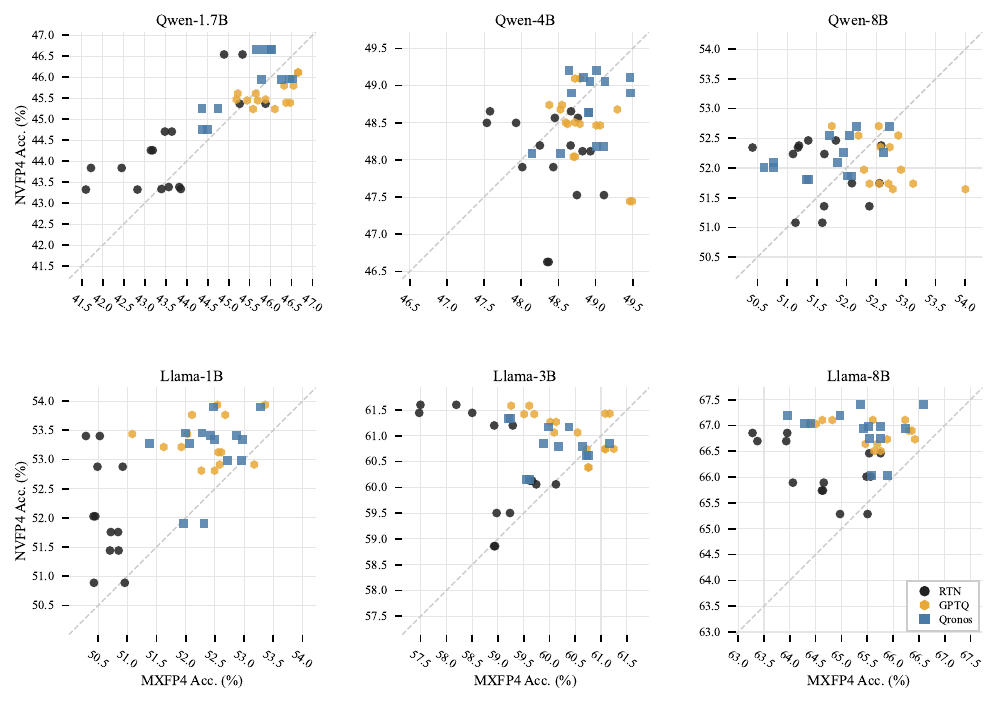}
    \caption{\textbf{PTQ Method Performance at A4W4.} Neither PTQ method is uniformly best by model or datatype.}
    \label{fig:ptq_method_all}
\end{figure}

Figure~\ref{fig:ptq_method_all} separates the A4W4 results by PTQ method. The relative ordering of GPTQ and Qronos changes across models and formats. GPTQ more frequently produces the strongest MXFP4 configurations, while Qronos more frequently produces the strongest NVFP4 configurations, but neither pattern holds universally. This interaction indicates that a PTQ method selected for one block format should not be assumed to transfer optimally to the other.

\section{Hadamard Performance by Model and PTQ Method}

\begin{figure}[h]
    \centering
    \includegraphics[width=\linewidth]{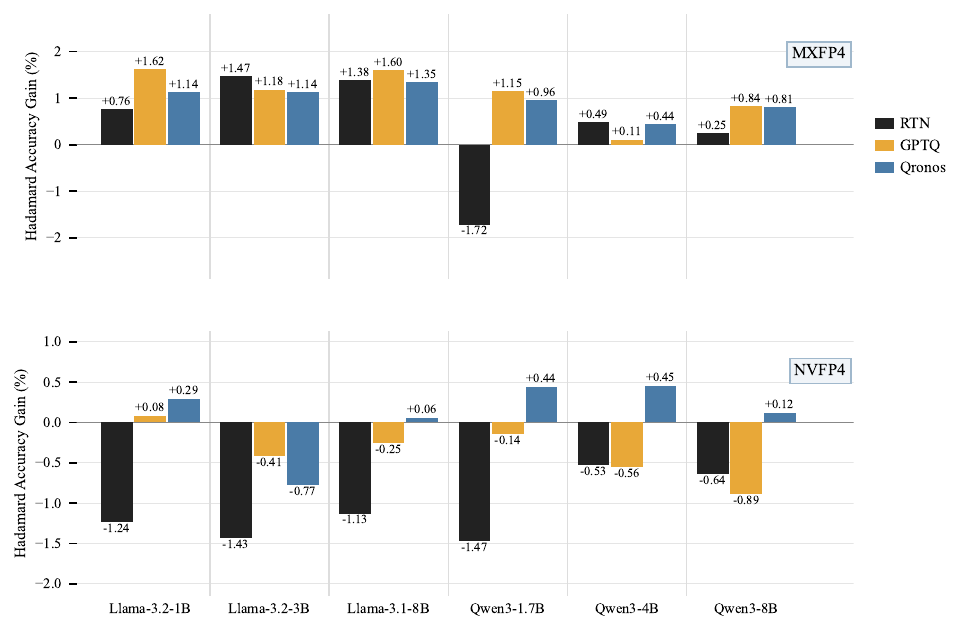}
    \caption{\textbf{Hadamard rotation $\times$ PTQ method.} We consider the average accuracy gain over no rotation for different Hadamard sizes (x-axis) on both NVFP4 (left) and MXFP4 (right) at A4W4. On average, Hadamard rotation does not have a positive effect on NVFP4, yet yields significant benefits for MXFP4. While these represent trends averaged over all  configurations, a best PTQ recipe for a given model may not follow this trend as shown in Table~\ref{tab:scale_config_sensitivity}.}
    \label{fig:hadamard_all}
\end{figure}

Figure~\ref{fig:hadamard_all} reports the average accuracy gain from Hadamard rotation relative to the corresponding configuration without rotation. For NVFP4, the mean effect is small or negative across rotation sizes. MXFP4 benefits more consistently, with Block-32 rotation producing the strongest average improvement. The gain also depends on the PTQ method, indicating that rotation and reconstruction-based quantization should be treated as interacting components of the PTQ recipe.

These aggregate trends do not imply that Block-32 rotation is optimal for every model. Individual configurations may favor no rotation, Block-16 rotation, or Full rotation. Table~\ref{tab:scale_config_sensitivity} places the rotation results in the context of the complete configuration sweep and reports the best and worst accuracy observed for each model and datatype.

\begin{table}[h]
\centering
\caption{There is no winning PTQ recipe across model and datatype, reinforcing the need for evaluating multiple PTQ configurations per model. 
The gap is the difference between the best and worst average accuracy across
configurations. Bold denotes the highest peak accuracy for each model.}
\label{tab:scale_config_sensitivity}
\small
\setlength{\tabcolsep}{4pt}
\begin{tabular}{llcccl}
\toprule
Model & Datatype & Best Acc. (\%) & Worst Acc. (\%) & Gap & Best Configuration \\
\midrule

\multirow{2}{*}{Qwen3-1.7B}
 & NVFP4           & \textbf{46.80} & 43.32 & 3.48 & Block-32 + Qronos + Stats \\
 & MXFP4 & 46.66          & 40.99 & 5.67 & Block-16 + GPTQ + MSE \\
\addlinespace

\multirow{2}{*}{Qwen3-4B}
 & NVFP4           & 49.32          & 46.63 & 2.69 & No rotation + Qronos + Stats \\
 & MXFP4 & \textbf{49.49} & 46.23 & 3.27 & Block-16 + GPTQ + Midmax \\
\addlinespace

\multirow{2}{*}{Qwen3-8B}
 & NVFP4           & 52.91          & 51.08 & 1.83 & Full + Qronos + Stats \\
 & MXFP4 & \textbf{54.00} & 49.54 & 4.46 & Full + GPTQ + MSE \\
\addlinespace

\multirow{2}{*}{Llama-3.2-1B}
 & NVFP4           & \textbf{53.94} & 50.89 & 3.05 & Block-16 + GPTQ + MSE \\
 & MXFP4 & 53.36          & 47.49 & 5.87 & Block-16 + GPTQ + MSE \\
\addlinespace

\multirow{2}{*}{Llama-3.2-3B}
 & NVFP4           & \textbf{61.68} & 58.70 & 2.98 & No rotation + Qronos + Stats \\
 & MXFP4 & 61.25          & 57.04 & 4.21 & Block-32 + GPTQ + Midmax\\
\addlinespace

\multirow{2}{*}{Llama-3.1-8B}
 & NVFP4           & \textbf{67.40} & 65.28 & 2.12 & Block-32 + Qronos + MSE \\
 & MXFP4 & 66.59          & 62.40 & 4.19 & Block-32 + Qronos + Midmax \\

\bottomrule
\end{tabular}
\end{table}

The configuration gap in Table~\ref{tab:scale_config_sensitivity} quantifies the importance of jointly tuning PTQ method, rotation, and scale selection. MXFP4 exhibits a larger best-to-worst gap than NVFP4 for every evaluated model, indicating greater sensitivity to the PTQ recipe. This sensitivity creates both greater risk from a poorly matched configuration and greater potential upside from format-specific tuning, as demonstrated by the strongest Qwen3-4B and Qwen3-8B results.

\end{document}